\documentclass[a4paper, 10pt, conference]{ieeeconf}      
\usepackage{FG2021}

\FGfinalcopy 

\IEEEoverridecommandlockouts                              
\usepackage{graphics} 
\usepackage{epsfig} 
\usepackage{mathptmx} 
\usepackage{times} 
\usepackage{amsmath} 
\usepackage{amssymb}  
\usepackage{multirow}

\def\FGPaperID{162} 

\title{\LARGE \bf
\textit{HeadPosr}: End-to-end Trainable Head Pose Estimation using Transformer Encoders
}

\author{\parbox{16cm}{\centering
    {\large Naina Dhingra}\\
    {\normalsize
    ETH Zurich\\
    ndhingra@ethz.ch}}
}

\begin{document}

\IEEEoverridecommandlockouts\pubid{\makebox[\columnwidth]{978-1-6654-3176-7/21/\$31.00~\copyright{}2021 IEEE \hfill} \hspace{\columnsep}\makebox[\columnwidth]{ }}

\ifFGfinal
\thispagestyle{empty}
\pagestyle{empty}
\else
\author{Anonymous FG2021 submission\\ Paper ID \FGPaperID \\}
\pagestyle{plain}
\fi
\maketitle

\begin{abstract}

 In this paper, \textit{HeadPosr} is proposed to predict the head poses using a single RGB image. \textit{HeadPosr} uses a novel architecture which includes a transformer encoder. In concrete, it consists of: (1) backbone; (2) connector; (3) transformer encoder; (4) prediction head. The significance of using a transformer encoder for HPE is studied. An extensive ablation study is performed on varying the (1) number of encoders; (2) number of heads; (3) different position embeddings; (4) different activations; (5) input channel size, in a transformer used in \textit{HeadPosr}. Further studies on using: (1) different backbones, (2) using different learning rates are also shown. The elaborated experiments and ablations studies are conducted using three different open-source widely used datasets for HPE, i.e., 300W-LP, AFLW2000, and BIWI datasets. Experiments illustrate that \textit{HeadPosr} outperforms all the state-of-art methods including both the landmark-free and the others based on using landmark or depth estimation on the AFLW2000 dataset and BIWI datasets when trained with 300W-LP. It also outperforms when averaging the results from the compared datasets, hence setting a benchmark for the problem of HPE, also demonstrating the effectiveness of using transformers over the state-of-the-art. 
	
\end{abstract}

	\section{Introduction}
	
	Facial expression detection, tracking, and modeling is an actively researched topic over the past years \cite{bulat2017far,kumar2017kepler}. There are several sub-problems of facial expression detection such as face recognition, age estimation, identification, HPE, etc. which are being analysed extensively by computer vision community. In this paper, we focus on HPE, which is an elemental step in computer vision used for identifying faces, localizing eye gaze, or approximating social interaction. It also aids in providing information in expression recognition, attention analysis, identity detection \cite{tran2017disentangled}, etc. It has several direct applications in various fields such as human-robot interaction, helping blind and visually impaired people in non-verbal communication \cite{dhingra2019res3atn}, assistance in driver behavior analysis, autonomous driving, etc.    
	
	\begin{figure}[tbp]
		\setlength\abovecaptionskip{-0\baselineskip}
		\begin{center}
			\includegraphics[height=6.5cm, width=\linewidth]{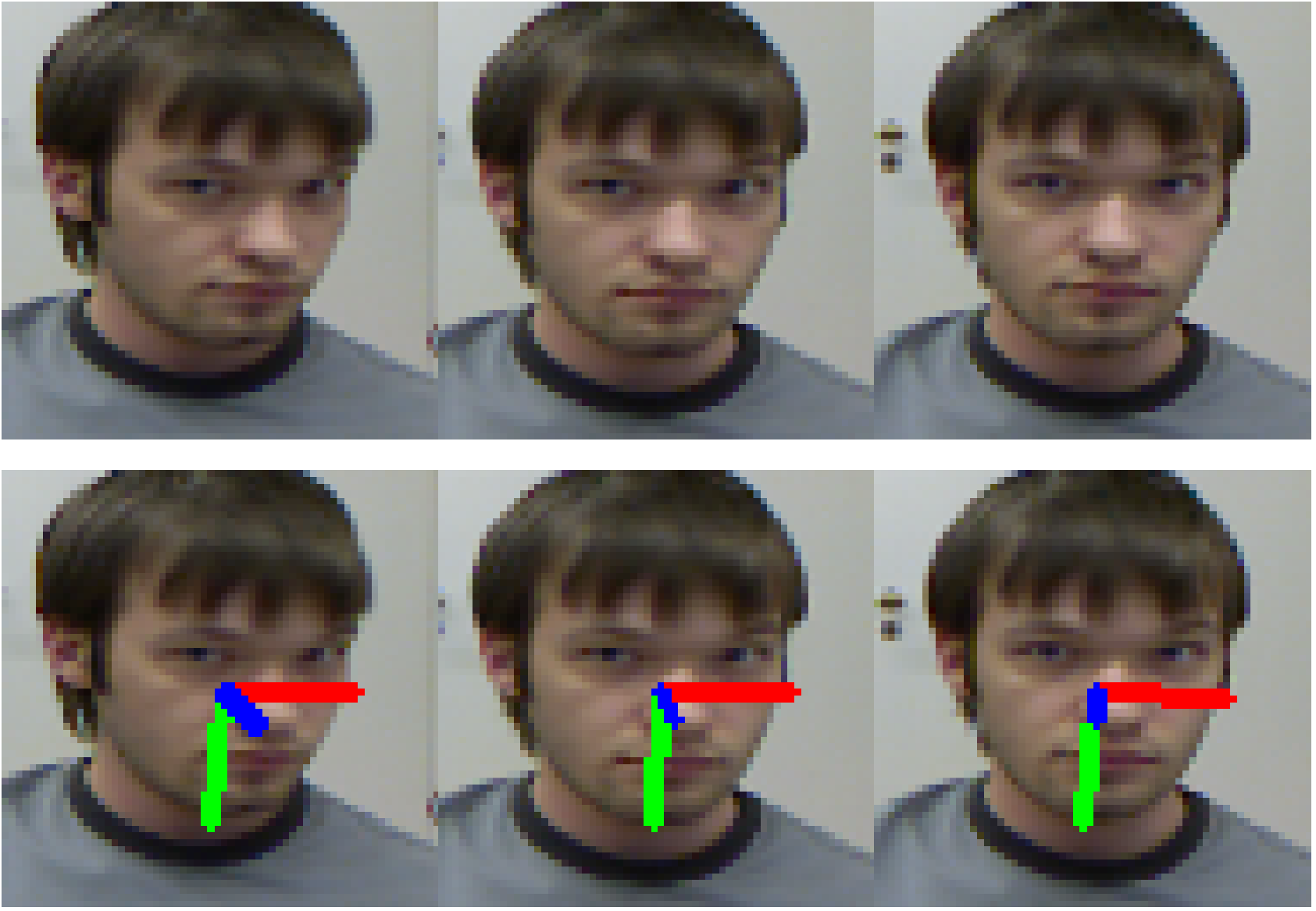}
			\caption{Sample image sequence taken from BIWI dataset. The corresponding head pose illustration is shown in the second row using ground truth head pose predictions. Different colored lines depict: Blue, shows the facing direction of the subject; Green shows downward direction; Red points towards the side.} 
			\label{fig:samplesequence}
		\end{center}
	\end{figure}
	
	HPE can be solved using facial keypoint detection. Some approaches \cite{bulat2017far,zhu2016face,kumar2017kepler} that use keypoints based detection illustrate robustness and flexibility in problems dealing with extreme pose changes and with varied visual scene occlusion (by matching correspondences in the images to their 3D face models). However, they face major challenges in the situations where it is difficult to detect landmarks. Moreover, these keypoints in literature are used for facial expression analysis. The head poses are generally a side result of the main task \cite{kumar2017kepler,ranjan2017hyperface} and are not very accurate on their own. Another solution is a holistic approach using image intensities to estimate the 3D head pose. This approach is demonstrated to perform better than the keypoints based one in \cite{ruiz2018fine}. The keypoints based approach has several drawbacks: (1) it requires good location of the keypoints, otherwise, the pose can not be estimated accurately; (2) the performance related to the estimation of head pose largely depends on the aspects of the 3D head models. Furthermore, the process of deforming the models is expensive and cumbersome as it has to be adapted to each participant. In some approaches, RGB-D images are used which provide good accuracy but have several drawbacks \cite{ruiz2018fine}. Data rates for RGB-D cameras are more than for RGB cameras which in turn require higher storage abilities and data transfer rates. Power requirements are also higher for RGB-D cameras. RGB-D cameras use active sensing which makes them difficult to be used in uncontrolled environments, especially outdoors. Due to these reasons, RGB-D based cameras cannot be used in applications such as mobile based devices, autonomous navigation, security purposes, computer graphics, etc. Consequently, in this work, a holistic based approach is used for estimating the head pose in RGB images. The image sequences with their ground truth head poses from BIWI dataset are illustrated in Figure~\ref{fig:samplesequence}.
	
	Inspired from the performance of transformer network \cite{vaswani2017attention}  in natural language processing field, many researchers in computer vision field \cite{dosovitskiy2020image} are exploring its application in image classification, image segmentation, image captioning, image registration, etc. Considering its positives for these applications, transformers are explored in this work for HPE. To our best knowledge, this is the first work to use transformer encoder to capture the spatial embedding from the head pose features for HPE.

	The main contributions of this paper are as follows:
	\begin{itemize}
		\item Proposing a method to predict head pose from the RGB images. For this, a novel architecture is developed based on a transformer encoder. It consists of four sub-parts: (1) backbone: to extract spatial image features; (2) connector: to downsample the output features from the backbone and then to further reshape them suitable input for a transformer encoder in the form of sequential data; (4) prediction head: to manipulate the output from the transformer encoder to predict the three head pose angles, i.e., yaw, pitch, roll.
		\item Comparing proposed \textit{HeadPosr} extensively with the state-of-the-art methods.
		\item Performed extensive ablation study for understanding the factors in estimating the head poses using \textit{HeadPosr}.
		\item  Qualitatively illustrating and comparing the results of the proposed approach to the ground truth, and a state-of-the-art.
		\item Outperforming published methods which use a single image for estimation of head poses when trained using 300W-LP and tested using combined AFLW2000 and BIWI datasets. It also performs better than all approaches when taken average of all the results. We also expect that our work will give further advances in head pose estimation using transformers.
		
	\end{itemize}

	This paper is structured as follows. Related work is discussed in section \ref{related_work}. The proposed transformer based network for HPE is described in section \ref{methodology}. The
	experiments are elaborately illustrated in section \ref{experiments}, results are stated in section \ref{results}, ablation study is discussed in section \ref{section:ablationstudy}. Finally, section \ref{summary} summarizes the presented work and prospects of future work.
	\section{Related Work} \label{related_work}
	
	\subsection{Head Pose Estimation}
	HPE can be grouped broadly in two categorizes: (1) the first category deals with detecting facial landmarks (keypoints or features) and adopting a reference 3D head model for corresponding feature matching; (2) the second category uses a complete face to recognize the head pose. They either use a appearance face model or they estimate the head pose directly from an image.  We will discuss literature study on these two categories as follows: 
	
	\subsubsection{Landmark Based Techniques}
	The head pose is estimated by detecting the correspondences between 2D image and 3D model landmarks. In \cite{dollar2010cascaded}, a cascaded pose regressor is proposed which refines the regression output at every refinement. Other regression based techniques are used in \cite{cao2014face,xiong2015global} which sketch rough faces to align them to real faces using regression. Some techniques detect the keypoints on faces using trained appearance models \cite{matthews2004active}. In \cite{sun2013deep}, a three level CNN is used to detect the position of landmarks on the face. In \cite{zhu2016face}, a cascaded neural network termed as 3D dense face alignment is proposed. This network fits a 3D morphable face model to an image. In \cite{guo2020towards}, a novel technique to optimize the regression task of estimating 3DMM parameters is used. It predicts the rotation matrix using a network. In \cite{kumar2017kepler}, a CNN based network is proposed to iteratively refine the position of facial keypoints. It also provides head pose information as a side result in terms of Euler angles. In \cite{ranjan2017hyperface}, a multi-task CNN model which can predict pose estimation, feature localization, gender recognition, and face detection from the same network is proposed. 
	
	The performance of all these techniques is conditioned on the landmark detection accuracy, hence due to these reasons we do not follow a landmark based approach. Also, it increases computation when pose can also be estimated without locating the keypoints.

	\subsubsection{Non Landmark Based Techniques}
	These techniques improved the state-of-the-art results on HPE. In \cite{fanelli2011real}, depth images and discriminative random regression forests are used for HPE. \cite{meyer2015robust} also uses depth images and registered 3D morphable models to these images. It refines the registration incrementally. In \cite{ruiz2018fine}, a CNN based network is proposed having multi-losses. This network uses an image to predict Euler angles and demonstrates that the HPE without landmarks performs better than with the landmarks. \cite{hsu2018quatnet} built a technique using quaternion that removes the gimbal lock issue which comes with Euler angles. \cite{yang2019fsa}, too uses a CNN based model to perform regression stage-wise. They use feature aggregation technique to group the spatial features along with attention mechanism. \cite{cao2021vector} uses rotation matrix with three vectors for representing the HPE and develops a CNN based architecture using the new representation. \cite{zhang2020fdn} proposes a feature decoupling network which learns the discriminative feature corresponding to each pose angle. It adopts a new cross-category loss for the network optimization. \cite{zhou2020whenet} proposes WHENet which uses multi-loss technique for training and this network is applicable for full range of head yaws. The better estimation of head pose using non-landmark techniques has induced this work to use this category of technique.
	
	
	\begin{figure*}[htbp!]
		\setlength\abovecaptionskip{-0\baselineskip}
		\setlength\belowcaptionskip{2pt}
		\begin{center}
			\includegraphics[height=8cm, width=17.5cm]{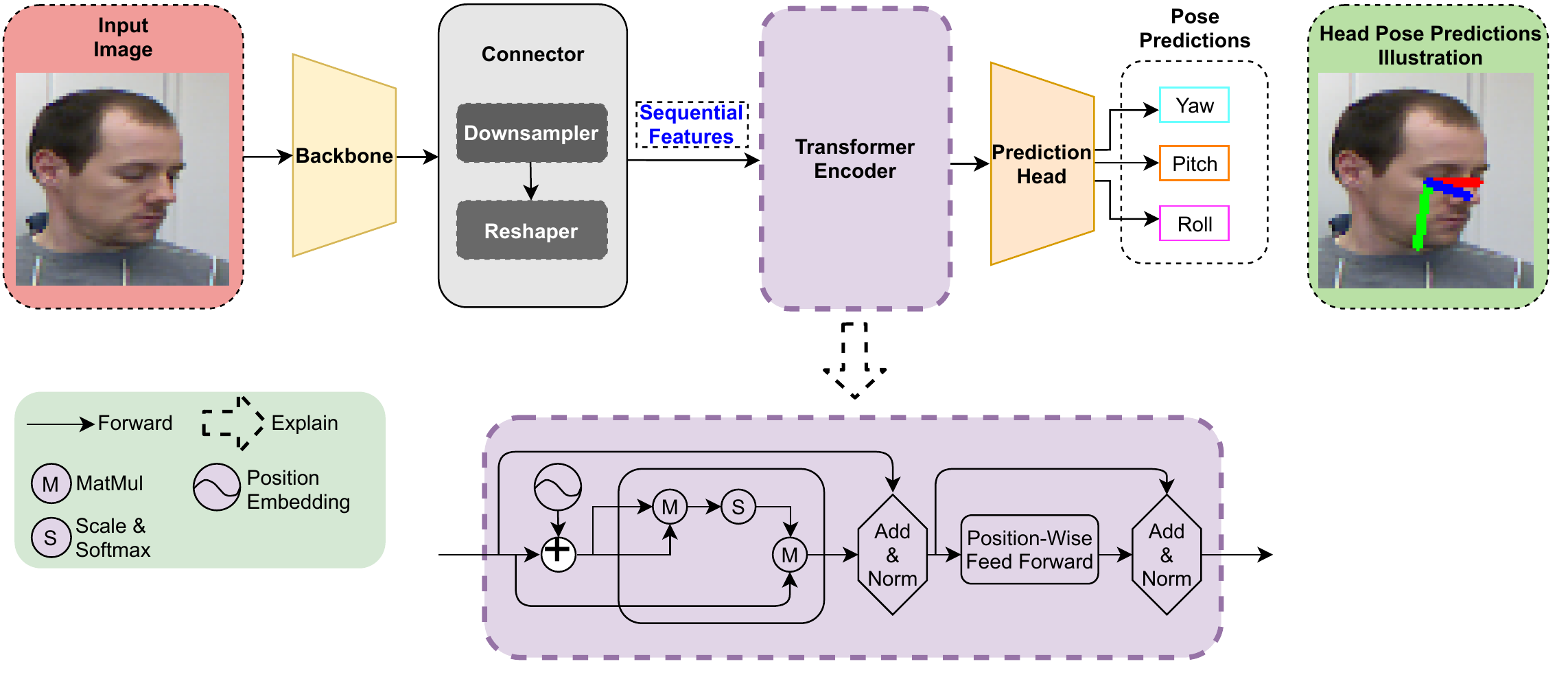}
			\caption{The framework of \textit{HeadPosr} is described in this figure. It consists of fours parts: (1) backbone: input RGB image is fed into backbone network. The output features are passed on to (2) connector which down-samples the input features and reshapes them to have a sequential form. These sequential features are used as input to a (3) transformer encoder. The output of the transformer encoder is fed into prediction head which outputs three vectors corresponding to the head pose, i.e., yaw, pith and roll. The whole framework is trained in end to end fashion. }
			\label{headposer}
		\end{center}
	\end{figure*}

	\subsection{Transformers}
	After achieving significant success in natural language processing \cite{vaswani2017attention}, transformers have gained a lot of attention among computer vision researchers. ViT \cite{dosovitskiy2020image} for image classification uses a transformer on sequences of image patches which outperforms CNNs. Various different versions of vision transformers have also been proposed recently: using a pyramid structure like CNNs \cite{wang2021pyramid}, performing data efficient training using distillation \cite{touvron2020training}, learning an abstract representation for self attention to improve the efficiency \cite{wu2021centroid}, manipulating it to have layer wise token to token transformation for encoding localized structure for individual token \cite{yuan2021tokens}. Vision transformers have gained popularity in tasks such as object segmentation, object detection, pose estimation, etc. The use of transformer for 3D human pose estimation \cite{li2021lifting} and 3D hand pose estimation \cite{huang2020hand} has motivated this work to use transformer for HPE. This is the first work as per our knowledge on HPE which employs transformers in the architecture. \\

	\section{Methodology} \label{methodology}
	
	\subsection{Problem Formulation}
	RGB image based HPE problem can be formulated as: Given N training images $I={i_n | n= 1, ...., N}$ that have pose vectors $v_n$ corresponding to each image $i_n$. This $v_n$ pose vector is a 3D vector, i.e., $v_n=[y, p, r]$, where $y$, $p$, $r$ corresponds to yaw, pitch, and roll angles, respectively. Thus, the aim of this problem is to find a relation $R$ such that $\bar{v}=R(i)$. This relation can predict the pose vector $\bar{v}$ as closely as possible to ground truth pose vector $v$. The relation $R$ is calculated using the optimization of the mean absolute error (MAE) between the ground truth and predicted poses.
	\begin{equation}
	(I)= \frac{1}{N}\sum_{n=1}^{N}|\bar{v}_n - v|
	\end{equation}
	
	where $\bar{v}_n=R(i_n)$ is the estimated pose vector from the proposed \textit{HeadPosr} network for the given image $i_n$. So, this formulates into a regression problem. Hence, this network needs to regress the pose vectors to solve the task of HPE. For this regression task, a transformer based architecture is built which employs CNN-based backbone to learn the head poses.

	\subsection{\textit{HeadPosr} Architecture}
	\textit{HeadPosr} network consists of a CNN based backbone, connector, transformer encoder, and head pose prediction head. Figure \ref{headposer}, shows the architecture of the network. The substructures of the network are described as follows:

	\subsubsection{Backbone}
	The backbone in the architecture consists of convolutional layers, very similar to those used in \cite{stoffl2021end,yang2020transpose}. Given a set of images, $I \in \mathbb{R}^{B \times C \times H \times W}$, $B$ stands for batch-size, $C$ stands for channels in the image, $H$ stands for height, $W$ stands for width of the image. This CNN with stride S generates lower-resolution feature maps $F \in \mathbb{R}^{B \times C' \times H/S \times W/S}$. In this work, different versions of ResNets \cite{he2016deep} with various strides S are used. To keep the network simple, we keep only the  initial layers of both the networks to extract features. 
	
	\subsubsection{Connector}
	Connector transform the output features from the backbone into a sequential data. It consists of two parts: (1) Downsampler, which is a $1 \times 1$ convolutional layer. It preserves the shape except the number of channels of the features. It is used as downsampler to reduce the dimension as well as to keep the condensed information. (2) Reshaper is used to change the shape of the features in a way so that the data is in sequential form. These output sequential features are used as an input to transformer encoder. The reshaper multiplies the height and width of the output features from the downsampler and then permutates them to transform into a sequential form. Considering the output set of features from the backbone as $F \in \mathbb{R}^{B \times C' \times H/S \times W/S}$, this output is channel downsampled as $F' \in \mathbb{R}^{B \times d \times H/S \times W/S'}$ where $d < C$. This $F'$ is flattened by reshaper in a way such that $F' \in \mathbb{R}^{B \times A \times d}$, where $A = H/S \times W/S$. This resulting formation of $F'$ is a sequential form and is fit to be an input to a transformer.
	
	\subsubsection{Transformer}
	In this architecture, we only use the encoder part of the transformer because regression is an encoding task and does not require a decoder to upscale the encoder output. The used transformer encoder architecture is a standard one \cite{vaswani2017attention}. This encoder compresses the sequential input to an encoded output. It passes through number of self-attention layers and feed-forward networks. Transformer encoder is able to understand the global information due to the self attention mechanism on the sequential information.\\
	\textit{Transformer Description:}
	Transformer works on the principle of multi-head self attention. This multi-head mechanism demands a sequential input ${F'} \in \mathbb{R}^{A \times d}$  to it. This is the reason that connector is used to convert the output from the backbone to a sequential input. The inputs are projected to $V  \in \mathbb{R}^{A \times d}$ values, $K  \in \mathbb{R}^{A \times d}$ keys, and $Q  \in \mathbb{R}^{A \times d}$ queries. They are calculated as:
	
	\begin{equation}
	Q_{i} = \hat{F'} * W^{Q}_{i}
	\end{equation}
	\vspace{-4mm}
	\begin{equation}
	K_{i} = \hat{F'} * W^{K}_{i}
	\end{equation}
	\vspace{-4mm}
	\begin{equation}
	V_{i} = \hat{F'} * W^{V}_{i}
	\end{equation}
	where $i$ stands for each attention head used. In the ablation study (Section \ref{section:ablationstudy}), the effect of varying the number of attention heads in \textit{HeadPosr} is shown.  $W^{Q}_{i} \in \mathbb{R}^{d \times d_{k}}$, $W^{K}_{i} \in \mathbb{R}^{d \times d_{k}}$ and $W^{V}_{i} \in \mathbb{R}^{d \times d_{v}}$ are parameter matrices.
	For detailed explanation of the transformers, refer to \cite{vaswani2017attention}. In this paper, similar to \cite{stoffl2021end,yang2020transpose}, position embedding is added instead of calculating the $V$. The transformer encoder without position embedding is a permutation-invariant which would disregard the spatial structure in the image. The effect of adding different types of positional embedding is also shown in ablation study (Section \ref{section:ablationstudy}). So, the resulting scaled dot-product attention head output $Z_{i}$ is formed as:
	\begin{equation}
	Z_{i} = softmax\left(\frac{Q_{i} * K_{i}^{T}}{\sqrt{d_{k}}}\right)
	\end{equation}
	
	The output $Z_{i}$ from each attention head is concatenated to get:
	\begin{equation}
	Z = concatenate(Z_{1}, Z_{2}, ... , Z_{n_{heads}})
	\end{equation}
	In this paper, effect of changing the number of attention heads is also studied in ablation study (Section \ref{section:ablationstudy}). The concatenated output acts as an input to a fully-connected layer. The output of the fully connected layer has same dimensions as the sequential input to the transformer encoder $\hat{F'} \in \mathbb{R}^{A \times d}$. The add and norm layer adds the input to the output of the multi-head attention as a residual connection. The normalization helps to deal with the `covariate shift' problem by recalculating the mean and variance of the cumulative inputs in every layer.

	\subsubsection{Head Pose Prediction Head}
	A very straightforward head pose prediction head is used in the network. Given the output from the transformer as  $F'_o \in \mathbb{R}^{B \times A \times C'}$, where $A = H/S \times W/S$, this head needs to predict ${R}^{B \times 3}$. Firstly, $F'_o \in \mathbb{R}^{B \times A \times C'}$ is reshaped to $F'_o \in \mathbb{R}^{B \times C' \times H/S \times W/S}$. If $ H/S$ and $ W/S$ are equal and smaller than 8, then the convolution layer with kernel size $ H/S \times W/S$ where $H=W$ is used. It also reduces the channel dimension from $A$ to 3 (corresponding to three head poses). In the ablation study (Section \ref{section:ablationstudy}), effect of using fully connected layer is also studied in comparison to convolutional layer.
	
	\section{Experiments} \label{experiments}
	\subsection{Datasets}
	
	\begin{figure}[hbtp!]
		\setlength\abovecaptionskip{-0\baselineskip}
		\setlength\belowcaptionskip{2pt}
		\begin{center}
			\includegraphics[height=5cm, width=\linewidth]{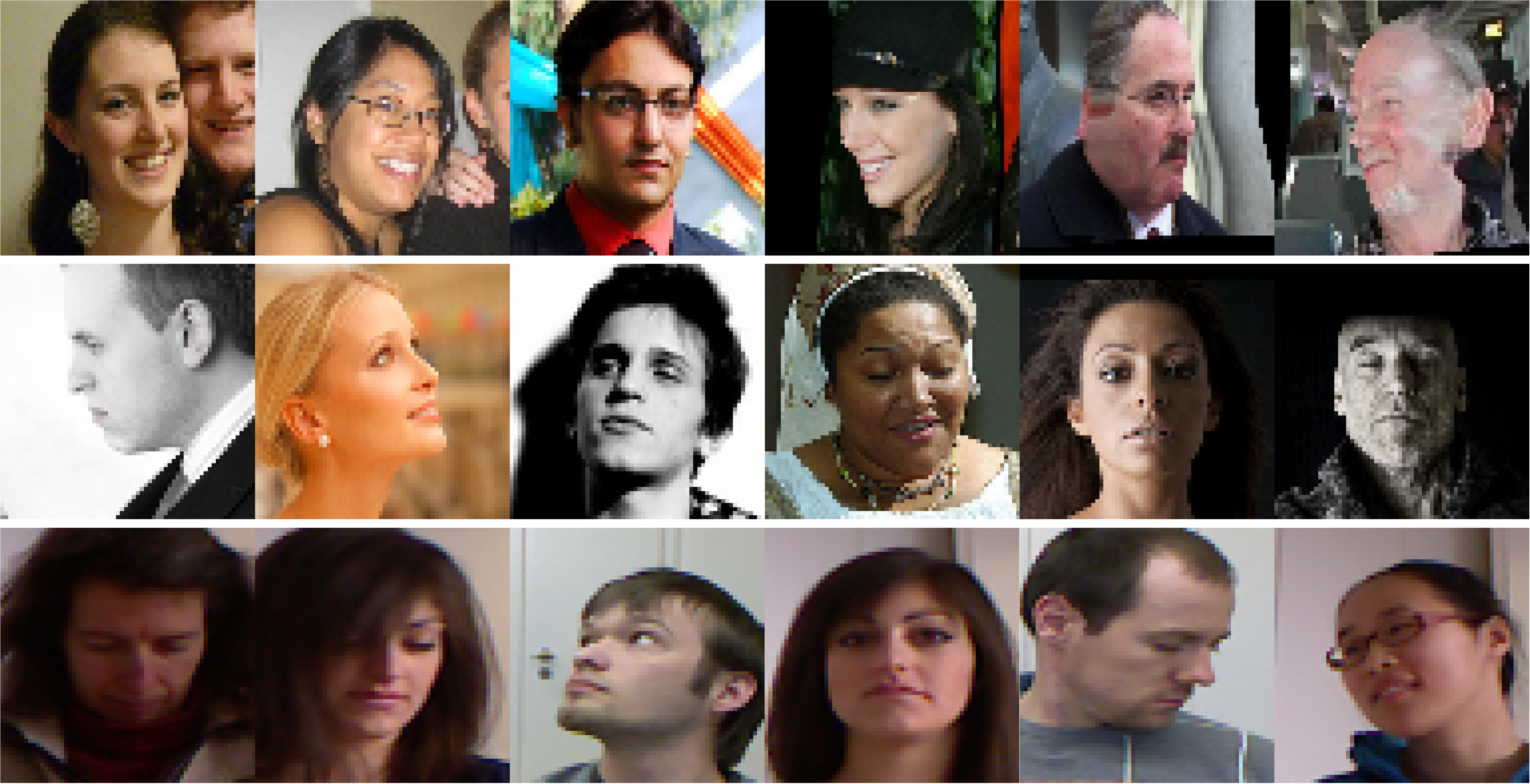}
			\caption{Some of the sample images from the datasets used for the experiments. Namely, first row:  300W-LP \cite{zhu2016face}; second row: AFLW2000 \cite{zhu2015high}; third row: BIWI \cite{fanelli2013random} dataset}
			\label{fig:datasetsamples}
		\end{center}
	\end{figure}
	
	The experiments were performed using three open-source datasets, i.e.,  300W-LP~\cite{zhu2016face}, AFLW2000~\cite{zhu2015high}, and BIWI~\cite{fanelli2013random} datasets. Some sample images from all the three datasets are shown in Figure \ref{fig:datasetsamples}.
	\subsubsection{300W-LP Dataset}
	This dataset is an extension of 300W dataset \cite{sagonas2013300} which is a combination of several datasets, for instance, HELEN \cite{zhou2013extensive}, AFW \cite{zhu2012face}, IBUG \cite{sagonas2013300}, LFPW \cite{belhumeur2013localizing}, etc. They used the technique of face profiling using 3D meshing for expanding this dataset to generate 61,225 image samples for large poses and used flipping to further expand to 122,450 synthesized image samples.
	
	\subsubsection{AFLW2000}
	This dataset has 2000 images from AFLW \cite{koestinger2011annotated} dataset. It has 3D faces with 68 landmarks corresponding to those faces. These faces have a wide range of illuminations, expressions and pose variations in its images.
	\subsubsection{BIWI}
	This dataset has total of 15,678 frames from 24 videos with 20 different subjects captured in the controlled indoor environment. This dataset does not include bounding boxes for the human heads. In this paper, similar to \cite{cao2021vector,yang2019fsa}, MTCNN \cite{zhang2016joint} is used to detect the human faces and crop the face to get the face bounding box output.
	
	This work uses the same protocols for training and testing as used by \cite{cao2021vector,yang2019fsa,doosti2020hope}, to have fair comparison with these approaches. We also filtered out the samples which have Euler angle outside the range of [-99\textdegree, 99\textdegree] as done by \cite{cao2021vector,yang2019fsa,doosti2020hope}. The two protocols used for the experiments are as follows.
	\begin{itemize}
		\item In this protocol, the model is trained with 300W-LP synthetic dataset and then the tests are conducted on the trained model by using AFLW2000 and BIWI datasets (both are real-world datasets), separately.
		\item In this set of experiments, training is performed by dividing the dataset into 70\% training and 30\% testing datasets as also done by \cite{cao2021vector,yang2019fsa}. These statistics results into 16 videos for training and 8 videos for testing from the 24 total videos. This protocol is used by several state-of-the-art work even when they used different types of modalities, i.e., RGB, RGB-D, or videos. In these paper, only RGB based data is used to perform the evaluation of our approach.
	\end{itemize}

	\begin{table*}[htbp!]
		\centering
		\caption{Comparison of the \textit{HeadPosr} with the state-of-the-art methods on BIWI\cite{fanelli2013random}(left) and AFLW2000\cite{zhu2016face}(right) dataset. The network is trained on 300W-LP dataset. (EH64 means 6 encoders, 4 heads in a transformer)}
		\label{Tab:result_300wlp}
		\begin{tabular}{l|llll|llll|l}
			\hline
			& \multicolumn{4}{c}{BIWI}   & \multicolumn{4}{|c}{AFLW2000} &  \multicolumn{1}{|c}{Total} \\
			\hline
			Method          & Yaw  & Pitch & Roll & MAE  & Yaw   & Pitch  & Roll & MAE & Total MAE \\
			\hline
			3DDFA~\cite{zhu2016face}            & 36.20 & 12.30 & 8.78  & 19.10 & 5.40  & 8.53  & 8.25  & 7.39 & 13.25 \\
			KEPLER~\cite{kumar2017kepler}     & 8.80  & 17.3 & 16.2 & 13.9 & -     & -     & -     & -     & 13.9\\
			Dlib (68 points)~\cite{kazemi2014one}    & 16.8 & 13.8 & 6.19  & 12.2 & 23.1 & 13.6 & 10.5 & 15.8 & 14 \\
			FAN (12 points)~\cite{bulat2017far}     & 6.36  & 12.3 & 8.71 & 9.12    & 8.53  & 7.48  & 7.63  & 7.88 & 8.5 \\
			Hopenet(a = 1)~\cite{ruiz2018fine}  & 4.81  & 6.61  & 3.27  & 4.90  & 6.92  & 6.64  & 5.67  & 6.41 &  5.65\\
			Hopenet(a = 2)~\cite{ruiz2018fine}  & 5.12  & 6.98  & 3.39  & 5.18  & 6.47  & 6.56  & 5.44  & 6.16 & 5.67\\
			Shao ~\cite{shao2019improving} & 4.59 & 7.25 & 6.15  & 6.00  & 5.07  & 6.37 & 4.99 & 5.48 & 5.74 \\
			SSR-Net-MD ~\cite{yang2018ssr}      & 4.49  & 6.31  & 3.61  & 4.65  & 5.14  & 7.09  & 5.89  & 6.01 & 5.33  \\
			FSA-Caps-Fusion~\cite{yang2019fsa} & 4.27  & 4.96  & 2.76  & 4.00  & 4.50  & 6.08  & 4.64  & 5.07  & 4.85  \\
			TriNet~\cite{cao2021vector} &       4.11 &  4.76 & 3.05  & 3.97 & \textbf{4.04}  & 5.78 & 4.20 & 4.67 & 4.32\\
			WHENet \cite{zhou2020whenet} & 3.99  & \textbf{4.39}  & 3.06  & 3.81 & 5.11  & 6.24  & 4.92  & 5.42 & 4.61 \\
			WHENet-V \cite{zhou2020whenet}        & 3.60  & 4.10  & 2.73  & \textbf{3.48}  & 4.44  & 5.75  & 4.31  & 4.83 & 4.16 \\
			
			\hline
			\textit{HeadPosr} EH38 & 4.08 & 5.10 &3.02& 4.06 & 4.60 & \textbf{4.86} & \textbf{2.87} & \textbf{4.11} & \textbf{4.08} \\
			\textit{HeadPosr} EH64 & \textbf{3.37} & 5.44 & \textbf{2.69} & 3.83 & 4.64 & 5.84 & 4.30 & 4.92 & 4.37 \\
		\end{tabular}
		
		\label{tab:narrow}
	\end{table*}
	
	For both the protocols, MAE results are reported for Euler angles.

	\subsection{Implementation Details}
	 \textit{HeadPosr} is implemented using Pytorch library. It employs the same data augmentation techniques as used by \cite{yang2019fsa,cao2021vector} to have the fair comparison. The random cropping and scaling (by a factor of 0.8 $\sim$ 1.2) was carried out in training. The network is trained for 90 epochs using an Adam optimizer with initial learning rate as 0.01. Further, this learning rate decayed by a factor of 0.1 after the interval of 30 epochs. Batch size used for training with 300W-LP dataset is 16 and with BIWI dataset is 8. The implementation was performed on  RTX 2080 Ti GPU.

	\section{Results} \label{results}

	\begin{table}[hbtp!]
		\caption{Comparison results of \textit{HeadPosr} with different methods on BIWI dataset. \textit{HeadPosr} is trained on BIWI dataset with 70\% training and 30\% testing data, using only RGB images.}
		\label{Tab:results_biwi}
		\centering
		
		\begin{tabular}{lllll}
			\hline
			& Yaw  & Pitch & Roll & MAE  \\
			\hline
			\multicolumn{5}{l}{\textbf{RGB-based}}                \\
			\hline
			DeepHeadPose~\cite{mukherjee2015deep}    & 5.67 & 5.18  & -    & -    \\
			SSR-Net-MD~\cite{yang2018ssr}      & 4.24 & 4.35  & 4.19 & 4.26 \\
			VGG16~\cite{gu2017dynamic}           & 3.91 & 4.03  & 3.03 & 3.66 \\
			FSA-Caps-Fusion~\cite{yang2019fsa} & 2.89 & 4.29  & 3.60 & 3.60 \\
			TriNet       &2.99 & 3.04 & 2.44 & 2.80 \\
			\textit{HeadPosr} EH64 & \textbf{2.59} &4.03 & 3.53 & 3.38 \\
			\hline
			\multicolumn{5}{l}{\textbf{RGB+Depth}}                \\
			\hline
			DeepHeadPose~\cite{mukherjee2015deep}    & 5.32 & 4.76  & -    & -    \\
			Martin~\cite{martin2014real}          & 3.6  & 2.5   & 2.6  & 2.9  \\
			\hline
			\multicolumn{5}{l}{\textbf{RGB+Time}}                 \\
			\hline
			VGG16+RNN~\cite{gu2017dynamic}       & 3.14 & 3.48  & 2.6  & 3.07
		\end{tabular}
	\end{table}

	\begin{figure*}[hbtp!]
		\setlength\abovecaptionskip{-0\baselineskip}
		\begin{center}
			\includegraphics[height=8.5cm, width=\linewidth]{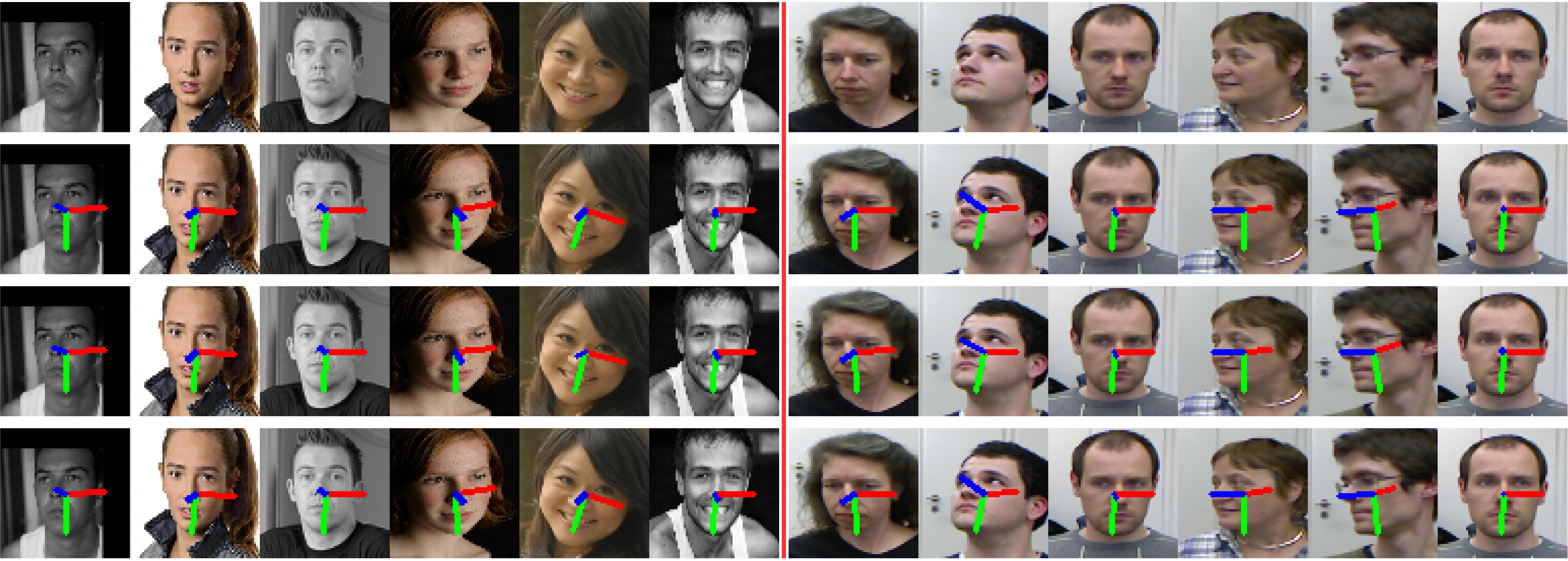}
			\caption{Qualitative Results of \textit{HeadPosr} in comparison to FSA-Net~\cite{yang2019fsa}. The red line after sixth column divides the figure in two parts: Left is from AFLW2000 dataset; Right is from BIWI dataset. The first row are test images from the datasets; second row is illustration of ground truth on the test image; third row is the output illustration from FSA-Net; forth row is the output illustration from the \textit{HeadPosr}}
			\label{fig:qualitativeresults}
		\end{center}
	\end{figure*}
	
	\textit{HeadPosr} is compared to state-of-art-methods as shown in Table \ref{Tab:result_300wlp} and \ref{Tab:results_biwi}. These compared networks are as follows. 3DDFA \cite{zhu2016face} utilizes CNNs to correspond the 3D model into an image. It is shown to work quite efficiently with occlusion scenes. KEPLER~\cite{kumar2017kepler} uses modified GoogleNet network to predict landmarks and pose simultaneously. Dlib~\cite{kazemi2014one} is a well known library which is used for landmark prediction, HPE, face detection, etc. FAN~\cite{bulat2017far} uses multi-scale features by merging features from multiple layers at multiple stages. Hopenet~\cite{ruiz2018fine} uses ResNet, similar to \textit{HeadPosr}, but is trained on mean squared error and cross entropy loss. Shao~\cite{shao2019improving} uses CNN based network by adjusting the margins of detected face bounding box. SSR-Net-MD~\cite{yang2018ssr}, FSA-Caps-Fusion~\cite{yang2019fsa}, and TriNet~\cite{cao2021vector} use the same basic technique of network having multiple stages using soft stage-wise regression. VGG16~\cite{gu2017dynamic} is a network having combination of CNN and RNN using Bayesian filters analysis. DeepHeadPose~\cite{mukherjee2015deep} uses low resolution depth images applying regression and classification to predict the pose. Martin~\cite{martin2014real} uses depth images for predicting head pose and uses registration between depth images and 3D model. WHENet~\cite{zhou2020whenet} uses efficientNet as backbone with multi-loss approaches and with changes to loss functions.\\

	For first set of experiments when \textit{HeadPosr} is trained with 300W-LP dataset, Table~\ref{Tab:result_300wlp} shows that it outperforms all the compared networks for the experiments trained using 300W-LP and hence setting it as a benchmark. It also performs quite closely on both the BIWI and AFLW2000 dataset.
	
	For the second set of experiment when \textit{HeadPosr} is trained with BIWI dataset, it outperforms all the approaches in predicting the yaw angle. It even surpasses the results from using different modalities. The MAE is more than that of TriNet. But it overall outperforms TriNet when taking average from both the sets of experiments.
	
	\subsection{Effectiveness of using a transformer}
	From the experimental study, it is evident that the using transformer encoder along with ResNet architecture performs quite well. Hopenet~\cite{ruiz2018fine} which is also a ResNet based architecture has MAE equivalent to 6.41 on AFLW2000 and 4.90  on BIWI dataset when trained using 300W-LP dataset. When \textit{HeadPosr} is compared to the performance of Hopenet, it can be seen in Table \ref{Tab:result_300wlp} that it outperforms AFLW2000 and BIWI datasets by ~36\%, and ~18\% in MAE respectively, when trained on 300W-LP dataset. These quantitative improvements also point out that transformers indeed are useful in employing for the task of HPE. There could be various ways of using them for this particular problem. This paper emphasizes that the transformers should be explored more for research in HPE.
	
	\subsection{Qualitative Examples}
	Figure \ref{fig:qualitativeresults} illustrates the HPE images from two datasets, namely, AFLW2000 and BIWI dataset. Both these datasets are used for testing when the network is trained with 300W-LP dataset. In this figure, the comparison between ground truth head pose illustration is made with the output from FSA-Net, and from \textit{HeadPosr}. It can be seen that \textit{HeadPosr} performs better than FSA-Net and is more closely related to the ground truth predictions illustration.

	\begin{table*}[hbtp!]
		\caption{Ablation Study results performed using \textit{HeadPosr}. The first set of ablations are on varying the parameters of the transformer encoder such as encoders, heads, activations, and position embeddings. The second set is on varying the learning rate, using different versions of prediction head, and dimension d of the input feature channel of transformer}
		\label{Tab:ablationtransformer}
		\centering
		\begin{tabular}{l|llll|llll|l}
			\hline
			& & BIWI &  &  &   & AFLW &  & & Total  \\
			\hline
			& Yaw  & Pitch & Roll & MAE & Yaw  & Pitch & Roll & MAE & Total MAE \\
			\hline
			\multicolumn{5}{l}{\textbf{No. of Encoder Layers}}                \\
			\hline
			1   & 4.75 & 4.92 & 2.84 & 4.17 & 4.47 & 5.68 & 4.09 & 4.74 & 4.45 \\
			2  & 4.70 & 4.87 & 2.72 & 4.10 & 4.27 & 5.84 & 4.20 & 4.77 & 4.43 \\
			3  &  4.16 & 5.24 & 2.93 & 4.11 & 4.34 & 5.06 & 4.25  & 4.55 & \textbf{4.33}\\
			4  & 4.40 & 4.90 & 2.78 & \textbf{4.02} & 4.38 & 5.77 & 4.15 & 4.76 & 4.39  \\
			5 & 4.45 & 4.93 & 2.94 & 4.10   & 4.33 & 5.74 & 4.23 & 4.76  & 4.43\\
			6 & 4.54 & 4.72 & 2.82 & \textbf{4.02} & 4.35 & 5.87 & 4.10 & \textbf{4.44} & 4.23 \\
			\hline
			\multicolumn{5}{l}{\textbf{No. of Heads}}                \\
			\hline
			1  & 4.81 & 4.82 & 2.86 & 4.16 & 4.34 &5.83 &4.13 & \textbf{4.76} & 4.46\\
			2 & 4.47 & 4.85 & 2.75 & 4.02 & 4.49 &5.85 & 4.27 & 4.87 & 4.45\\
			4 & 3.37 & 5.44 & 2.69 & \textbf{3.83} & 4.64 & 5.84 & 4.30 & 4.92 & \textbf{4.38}\\
			8 & 4.90& 4.41 & 2.87 & 4.06 & 4.47 & 5.81 & 4.26 & 4.84 & 4.45 \\
			16 & 3.78 & 5.34 & 2.90 & 4.01& 4.41 & 6.07 & 4.23 & 4.90 & 4.46\\
			\hline
			\multicolumn{5}{l}{\textbf{Activations}}                 \\
			\hline
			Relu &  4.08 & 5.10 &3.02& \textbf{4.06 }& 4.60 & 4.86 & 2.87 & \textbf{4.11} & \textbf{4.08}\\
			Gelu & 4.16 & 5.24 & 2.93 & 4.11 & 4.44 & 6.06 & 4.35  & 4.95 & 4.53\\
			\hline
			\multicolumn{5}{l}{\textbf{Position Embeddings}}                 \\
			\hline 
			No Embedding  &4.41 & 4.87 & 2.77 & \textbf{4.01} & 4.59 & 5.94 & 4.36 & 4.96 & 4.48\\
			Learnable &4.90 & 4.41 & 2.87 & 4.06 & 4.47 & 5.81 & 4.26 & \textbf{4.84} & \textbf{4.45} \\
			Sine  & 4.16 & 5.24 & 2.93 & 4.11 & 4.44 & 6.06 & 4.35 & 4.95 & 4.53\\
			\hline
			\hline
			\multicolumn{5}{l}{\textbf{Learning Rate}}                 \\
			\hline
			0.01 & 12.09 & 13.94 & 10.67 & 12.23 & 12.39 & 12.89 & 13.90 & 13.06 & 12.26\\
			0.001 & 4.08 & 5.10 &3.02& 4.06 & 4.60 & 4.86 & 2.87 & \textbf{4.11} & \textbf{4.08}\\
			0.0001 & 4.72 & 4.57 & 2.69 & \textbf{3.99} & 4.34 & 5.89 & 4.36 & 4.86 & 4.42\\
			\hline
			\multicolumn{5}{l}{\textbf{Types of Prediction Head}}                 \\
			\hline
			Conv $1 \times 1$ & 4.58 & 4.60 & 2.86 & \textbf{4.01} & 4.37 & 4.70 & 3.83 & 4.30 & 4.15\\
			Conv $8 \times 8$ &4.08 & 5.10 &3.02& 4.06 & 4.60 & 4.86 & 2.87 & \textbf{4.11} & \textbf{4.08}\\
			FC layer  & 4.32 & 4.73 & 3.01 & \textbf{4.01} & 4.25 & 4.82 & 4.20 & 4.42 & 4.21\\
			\hline
			\multicolumn{5}{l}{\textbf{Value of d}}                 \\
			\hline
			16  &4.08 & 5.10 &3.02&\textbf{4.06} & 4.60 & 4.86 & 2.87 & \textbf{4.11} &\textbf{4.08}\\
			32 & 4.23 & 4.89 & 4.99 & 4.70 & 4.74 & 4.62 & 4.01 & 4.45 & 4.57\\
			\hline
		\end{tabular}
		\vspace{1mm}
	\end{table*}

	\begin{table}[hbtp!]
		\caption{Ablation study using different backbones for comparing results on RGB images from BIWI dataset (training performed using BIWI dataset). Three different backbones are evaluated in the \textit{HeadPosr}, i.e., ResNet18, ResNet32, and ResNet50. The evaluation is performed using MAE.}
		\label{Tab:Resnets}
		\centering
		\begin{tabular}{lllll}
			\hline
			& Yaw  & Pitch & Roll & MAE  \\
			\hline
			\multicolumn{5}{l}{\textbf{Networks}}                \\
			\hline
			ResNet18       & 3.20 & 4.98 & 3.85 & 4.01 \\
			ResNet32        & 3.09 & 4.94 & 3.54 & 3.85   \\
			ResNet50       & \textbf{2.59} &\textbf{4.03} & \textbf{3.53} & \textbf{3.38} \\
			\hline
		\end{tabular}
	\end{table}

	\section{Ablation Study} \label{section:ablationstudy}
	The ablation study is conducted to understand the influence of various parameters in the \textit{HeadPosr}. For each comparison of the study to have fair comparison, we use same parameters for that study except the parameter being studied. 
	\subsection{Transformer}
	Several parameters of the transformer encoder were varied to check their effect on the HPE.
	\subsubsection{Number of Encoders} The number of encoder layers used for transformer encoder are varied as shown in Table \ref{Tab:ablationtransformer}. It is seen that three encoders performs better than others when MAE is averaged on both the testing datasets.
	\subsubsection{Number of Heads} The number of attention heads in the transformer encoder were varied from 1 to 16 by multiples of 2 to check their effects. The quantitative values are described in Table \ref{Tab:ablationtransformer}. It can be seen that the number of heads equal to 4 has better quantitative results than other compared number of heads (taking average of both the datasets).
	
	\subsubsection{Type of activation} In this work, two types of activation function in the transformer encoder were tested, i.e., Relu and Gelu. It is seen that Relu performed better than Gelu in \textit{HeadPosr} as shown in Table \ref{Tab:ablationtransformer}.
	
	\subsubsection{Position Embedding} Three types of position embedding are used to perform the experiments: (1) no embedding; (2) sine embedding; and (3) learnable embedding by adding a parameter. Learnable embedding performs better than the other two compared embeddings (see Table~\ref{Tab:ablationtransformer}).
	
	\subsection{Type of Backbone} Different versions of ResNet are evaluated such as: (1) ResNet18; (2) ResNet34; and (3) ResNet50. This evaluation is performed using BIWI dataset (as in protocol 2) for training and testing. It is seen in the evaluation as shown in Table \ref{Tab:Resnets} that ResNet50 works better than the other two compared networks as a backbone in the \textit{HeadPosr}. In our experiments, we use ResNet50.
	
	\subsection{Type of Prediction Head} Prediction head is used after transformer encoder and at the end of the network. It converts the output features into three output angles. Three types of prediction heads are studied: (1) fully connected layer; (2) convolutional layer with kernel size equal to length and width of the features (considering length = width $\leq$ 8); (3) convolutional layer with kernel size 1 and summation of features length and width axis. The experimental results are shown in Table \ref{Tab:ablationtransformer}. Considering the results, for further experiments, convolutional layer with kernel size 8 is used which has features having $8 \times 8$ length and width.
	\subsection{Channels of Input Features to Transformer} The input to transformer has $d$ channels (output from the connector). In the experiments, two dimensions of channels were studied, i.e. 16 and 32. From the table, it is evident that 16 performed better than 32 channels for the \textit{HeadPosr}.

	\section{Conclusion and Future Work} \label{summary}
	In this paper, \textit{HeadPosr}, a novel technique for HPE is proposed that can predict head poses using a transformer. This is achieved by using a network having a CNN based backbone, a connector for manipulating the features to be used as input for the transformer, a transformer encoder, and prediction head. \textit{HeadPosr} outperforms the already existing techniques when trained using 300W-LP. Overall taking average from BIWI and AFLW2000 datasets, it exceeds the performance of the state-of-the-art methods. To our knowledge, it is the first approach of using transformer for HPE. Hence, it paves a way to consider them for wider research in the problem of HPE. In the future, it is intended to extend this work by reducing the network size, exploring different transformer structures, and different loss function techniques. Further expansion would be to consider the modification of representation of head pose using a transformer.

{\small
\bibliographystyle{ieee}
\bibliography{main}
}

\end{document}